\documentclass[12pt]{article}

\usepackage[utf8]{inputenc}
\usepackage[T1]{fontenc}
\usepackage{amsmath,amssymb}
\usepackage{graphicx}
\usepackage{booktabs}
\usepackage{multirow}
\usepackage{hyperref}
\usepackage{xcolor}
\usepackage{authblk}
\usepackage{geometry}
\usepackage{caption}
\usepackage{subcaption}
\usepackage{longtable}
\usepackage{array}
\usepackage{tabularx}
\usepackage{float}
\usepackage{enumitem}
\hypersetup{
  colorlinks=true,
  linkcolor=blue,
  citecolor=blue,
  urlcolor=blue
}

\title{Explainable Machine Learning for Sepsis Outcome Prediction Using a Novel Romanian Electronic Health Record Dataset}

\author[1,\thanks{Equal contribution}]{Florentina Mu\c{s}at}
\author[2,\thanks{Equal contribution; Corresponding author: \href{mailto:andrei.bunea@upb.ro}{andrei.bunea@upb.ro}}]{Andrei-Alexandru Bunea}
\author[2]{Ovidiu-C\u{a}t\u{a}lin Ghibea}
\author[2]{Dan-Matei Popovici}
\author[1,3]{Ion Daniel}
\author[1,4]{Octavian Andronic}

\affil[1]{Carol Davila University of Medicine and Pharmacy, Faculty of Medicine, Bucharest, Romania}
\affil[2]{POLITEHNICA Bucharest National University of Science and Technology, Bucharest, Romania}
\affil[3]{General Surgery Department, University Emergency Hospital of Bucharest, Bucharest, Romania}
\affil[4]{Innovation and eHealth Center, Carol Davila University of Medicine and Pharmacy, Bucharest, Romania}

\date{}

\begin{document}

\maketitle

\begin{abstract}
We develop and analyze explainable machine learning (ML) models for sepsis outcome prediction using a novel Electronic Health Record (EHR) dataset from 12,286 hospitalizations at a large emergency hospital in Romania. The dataset includes demographics, International Classification of Diseases (ICD-10) diagnostics, and 600 types of laboratory tests. This study aims to identify clinically strong predictors while achieving state-of-the-art results across three classification tasks: (1)~deceased vs.\ discharged, (2)~deceased vs.\ recovered, and (3)~recovered vs.\ ameliorated. We trained five ML models to capture complex distributions while preserving clinical interpretability. Experiments explored the trade-off between feature richness and patient coverage, using subsets of the 10--50 most frequent laboratory tests. Model performance was evaluated using accuracy and area under the curve (AUC), and explainability was assessed using SHapley Additive exPlanations (SHAP). The highest performance was obtained for the deceased vs.\ recovered case study (AUC\,=\,0.983, accuracy\,=\,0.93). SHAP analysis identified several strong predictors such as cardiovascular comorbidities, urea levels, aspartate aminotransferase, platelet count, and eosinophil percentage. Eosinopenia emerged as a top predictor, highlighting its value as an underutilized marker that is not included in current assessment standards, while the high performance suggests the applicability of these models in clinical settings.

\end{abstract}

\section{Introduction}

Sepsis remains one of the leading causes of morbidity and mortality in hospitals worldwide, posing a significant clinical and operational challenge for healthcare systems. Characterized by a dysregulated host response to infection leading to life-threatening organ dysfunction, the early detection and accurate prognosis of sepsis are crucial for improving patient outcomes and optimizing resource allocation.

Sepsis is a global health burden~\cite{rudd2020}, yet sepsis-related mortality varies markedly across regions, with lower age-standardized mortality rates in 2017 reported in Western European countries (25.7 per 100,000 individuals)~\cite{acsqhc2021}, compared to Central (42.3 per 100,000 individuals) and Eastern European countries (64.6 per 100,000 individuals), including Romania (54.1 per 100,000 individuals)~\cite{musat2025sepsis}.

Some of the most widely utilized scoring systems for establishing prognosis in sepsis are the Sequential Organ Failure Assessment score (SOFA)~\cite{vincent1996}, the quick sepsis-related organ failure assessment (qSOFA)~\cite{rather2015}, the acute physiology and chronic health evaluation (APACHE)~\cite{donabedian2002}, the simplified acute physiology score (SAPS)~\cite{legall1993}, among others. Although these scores aim to provide an objective and standardized way in assessing the septic state, they also present significant limitations---they rely on a predefined set of variables and assume linear relationships between lab test results and patient outcomes. These systems act as a checklist with limited generalizability across diverse clinical settings and population cohorts~\cite{vincent1996}. Most importantly, they are unable to capture the complex, non-linear associations between underlying factors that might catch an early aggravation in the septic state~\cite{wongtangman2021}.

Critical care datasets coming from the United States are extensive, with databases like eICU-CRD containing even larger patient pools ($\sim$200,000 ICU stays across hospitals), but cover shorter time spans (e.g., 2014--2015) and mix patient populations~\cite{pollard2018}. In contrast, our cohort gathering around 12,000 cases from a single tertiary hospital is one of the largest from a developing European healthcare system, and its 18-year span offers a long-term robust perspective.

Centralized records on sepsis have been scarce in Central and Eastern Europe, forcing researchers to rely on foreign data to perform comprehensive studies, or rely on smaller cohorts. In contrast, our Romanian cohort captures detailed EHR data with up to 600 lab tests, demographics and diagnoses, allowing the training of feature-rich models.

Our previous study emphasizes that sepsis is a deadly threat in Romanian hospitals, and outcomes lag behind those in wealthier countries, concluding that early recognition and improved allocation of ICU resources are the principal means of improving the septic outcome~\cite{musat2025sepsis}. We develop further our previous work, and we tackle in this study this precise gap that our ML solution aims to fill.

We introduce a novel, machine learning--driven approach to address these challenges. Specifically, we train 5 predictive models to estimate three distinct outcomes associated with sepsis progression: \textbf{clinical improvement}, \textbf{complete recovery}, and \textbf{mortality}. In contrast to prior studies on sepsis prediction, our results demonstrate superior performance on the most critical task of Deceased vs.\ Discharged patient outcome, achieving state-of-the-art accuracy compared to existing methods.

Our work relies on a large Romanian sepsis cohort, gathering over 12,000 adult hospitalizations spanning an 18-year period (2007--2024) at one of the largest emergency hospitals in the region---The University Emergency Hospital of Bucharest (UEHB). To the authors' knowledge, this is the first sepsis outcome prediction study using Machine Learning in an Eastern European country. Most ML-driven efforts for sepsis mortality prediction have been developed on North American (MIMIC-III, MIMIC-IV, eICU-CRD) and Western European datasets.

Learning complex distributions over different hospitalization outcomes (deceased vs.\ recovered, deceased vs.\ ameliorated, ameliorated vs.\ worsened) and complex patient profiles (from 43 ICU units, and 600 investigation types) we train our models to detect subtle predictors, potentially outperforming generic scores. Going forward, combining such data-driven insights with clinical expertise could help bridge the outcome disparity between Romania and more developed countries.

Although advanced intensive care units (ICUs) enable comprehensive patient monitoring, many healthcare facilities---particularly those with limited resources---depend primarily on basic diagnostic capabilities. The broad coverage of our dataset across the Romanian population, and potentially the wider Eastern European demographic, together with the reliance solely on standard routine clinical laboratory tests, makes our approach readily deployable across diverse hospital settings.

\section{Materials and Methods}

\subsection{Dataset}

This study utilized electronic medical data obtained from The University Emergency Hospital of Bucharest (UEHB), Romania's largest emergency care institution, accommodating 1,099 beds and treating over 49,000 patients in 2024 alone.

We retrospectively identified adult patients diagnosed with sepsis who were admitted to UEHB between January 1, 2007, and December 31, 2024. Patient selection was based on a curated list of ICD-10 diagnostic codes indicative of sepsis (details available in Supplementary Table~S1). Inclusion criteria extended to individuals with either a primary or secondary sepsis-related ICD-10 code, as well as those suspected or diagnosed with sepsis according to clinicians' manual entries in the electronic medical records. These are non-standardized descriptions such as ``urosepsis'' or ``septic shock due to pneumonia''---which can often provide clinical context not fully captured by codified diagnoses.

Exclusion criteria included patients under 18 years of age, neonatology admissions, and hospital stays limited to a single day. Altogether, 12,089 patient records were extracted from the hospital's digital system, corresponding to the 12,286 hospitalizations studied in this work. In total, the study tracked 600 different types of laboratory tests and each patient record included the nature of the discharge (deceased or alive) and the type of discharge (deceased, worsened, improved, or recovered).

The study was conducted in accordance with the ethical principles outlined in the Declaration of Helsinki (1975), with its most recent revision in 2013. Ethical approval was granted by both the Ethics Committee of the Carol Davila University of Medicine and Pharmacy (Approval No.\ 6098/21.03.2025) and the University Emergency Hospital of Bucharest (Approval No.\ 36964/17.06.2022). Informed consent was waived by the Ethics Committee of the University Emergency Hospital of Bucharest due to the retrospective nature of the study and the use of anonymized data.

\subsection{Data Curation and Preprocessing}

Demographic and outcome details (sex, age, type of discharge, diagnostics) were linked with the laboratory investigations using unique patient identifiers, thus creating a patient-centric data structure.

\begin{figure}[ht]
\centering
\includegraphics[width=0.95\textwidth]{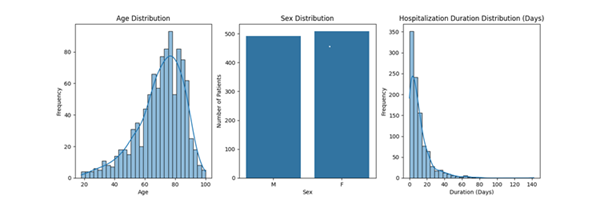}
\caption{Descriptive statistics of the dataset: (a) age distribution, (b) sex distribution, and (c) hospitalization duration.}
\label{fig:descriptive}
\end{figure}

To eliminate potential bias we conducted statistical insight on the distribution of the features that can cause prediction biases: sex, age, discharge outcome, and type of discharge. Figure~\ref{fig:descriptive} captures the distribution of these features, showing a normal distribution with most patients around 77 years old, a balanced sex distribution and a highly right-skewed distribution, with most of the patients being discharged before 25 days.

The majority of patients were deceased at discharge, while the remaining cases were discharged in various conditions (improved, recovered, or worsened) (Supplementary Figure~S1). Discharged patients that are not fully recovered were either discharged at their own request or transferred to another facility. While the dataset is already a balanced input for Deceased vs.\ Discharged classification, for predictions on discharge type such as Deceased vs.\ Recovered, we addressed the excessive imbalance by random undersampling (we randomly selected an equal number of patients from the majority class to match the number of cases in the minority class).

As the study gathers records from 43 ICU units, the laboratory tests in our dataset covered a wide range of physiological markers beyond sepsis-specific parameters. Collecting results from 600 investigation types, the data presents extreme sparsity that we addressed by experimenting with different subsets of features to identify the best trade-off between feature breadth and patient coverage. Thus, we created a set of 50 investigations collecting not only the tests referenced in medical-related work, but also tests that are frequently performed on septic patients but not yet established as highly relevant for the SOFA assessment (Supplementary Table~S2). To evaluate the trade-off mentioned above, we conducted experiments using the top 10, 20, 30, 40, and 50 most frequent investigations.

For each subset, we identified the hospitalization records with complete coverage. While 92.59\% of the patients took all the investigations from the top-10 subset, only 8.68\% of them have registered results for all top-50 most frequent tests (Table~\ref{tab:coverage}).

\begin{table}[ht]
\centering
\caption{Coverage of patients from the dataset by subset size of most frequent laboratory tests.}
\label{tab:coverage}
\begin{tabular}{lccccc}
\toprule
\textbf{Subset size} & Top-10 & Top-20 & Top-30 & Top-40 & Top-50 \\
\midrule
\textbf{\% of patients with full coverage} & 92.59\% & 72.40\% & 45.10\% & 23.99\% & 8.68\% \\
\bottomrule
\end{tabular}
\end{table}

Sepsis is a time-sensitive condition; however, due to data sparsity in this study, temporal variations in laboratory results were not considered. After evaluating multiple strategies for aggregating laboratory time series, we found that using the mean value over the hospitalization period yielded the best performance across all predictive tasks.

Variations of our models also include information about primary and secondary diagnostics. There are approximately 17,000 diagnostic codes in the ICD-10 standard, maintained by the World Health Organization (WHO). To incorporate them into our model without confronting excessive sparsity, we designed a simplified encoding mechanism, classifying the numerous ICD-10 diagnostic codes into 14 clinically relevant comorbidity categories presented in Supplementary Table~S1. We then appended this information to the input using one-hot encoding.

\subsection{Classification Scenarios and Experimental Framework}

We formulated and evaluated three binary classification scenarios to analyze sepsis outcomes, each reflecting a distinct clinical question:

\begin{enumerate}[leftmargin=*]
  \item \textbf{Deceased vs.\ Discharged (Alive):} This experiment provides the most important insight in clinical practice, predicting the chances of survival during the hospitalization period. The classification groups all non-deceased statuses (e.g., ameliorated, worsened, recovered) as a single category.

  \item \textbf{Deceased vs.\ Recovered:} This comparison covers the most extreme outcomes of hospitalization and distinguishes between patients who died of sepsis infection and the ones who fully recovered.

  \item \textbf{Recovered vs.\ Improved:} Focused exclusively on surviving patients, this comparison evaluates which individuals achieved full recovery versus the ones discharged in an ameliorated condition. It provides insights on the functional recovery among survivors, setting clear expectations if the patient is going to be completely cured.
\end{enumerate}

\section{Machine Learning Models}

We trained and evaluated five widely used machine learning algorithms on our classification tasks: Logistic Regression (LR), Support Vector Classifier (SVC), Random Forest (RF), Gradient Boosting (GB), and Histogram-based Gradient Boosting (HistGB).

Models were trained on five sets of laboratory tests: the Top-10, 20, 30, 40, and 50 most frequent investigations. Each of the five subsets was trained on two input configurations:

\begin{enumerate}[leftmargin=*]
  \item \textbf{Base variant:} Demographics + lab results
  \item \textbf{Extended variant:} Demographics + lab results + comorbidities
\end{enumerate}

This allowed us to measure the added value of comorbidity information in outcome prediction.

Model performance was assessed using accuracy and area under the receiver operating characteristic curve (AUC). Accuracy captures the overall correctness, while AUC evaluated the classification ability to create a clear distinction between patients that are likely to recover from those who are at high risk.

\section{Results}

Overall, the models trained with diagnostic features outperformed the ones with just the EHR data across all tasks. Furthermore, the top-40 subset of investigations demonstrated the best trade-off between a feature-rich dataset and patient coverage. One exception is the Deceased vs.\ Recovered task, where the top-20 reached the best accuracy, with the top-40 subset ranking second, just behind.

A summary of the best performing models for each classification task is given in Table~\ref{tab:best_results}.

\begin{table}[ht]
\centering
\caption{Best model results across all three classification tasks.}
\label{tab:best_results}
\footnotesize
\setlength{\tabcolsep}{4pt}
\begin{tabular}{llclccccc}
\toprule
\textbf{Case study} & \textbf{Variant} & \textbf{Subset} & \textbf{Best Model} & \textbf{Acc.} & \textbf{Prec.} & \textbf{Rec.} & \textbf{F1} & \textbf{AUC} \\
\midrule
Deceased vs.\ Disch. & diag & top-40 & HistGB & 0.843 & 0.843 & 0.843 & 0.843 & 0.920 \\
Deceased vs.\ Disch. & w/o diag & top-40 & GB & 0.794 & 0.794 & 0.794 & 0.794 & 0.866 \\
\addlinespace
Deceased vs.\ Recov. & diag & top-40 & HistGB & 0.930 & 0.932 & 0.930 & 0.930 & 0.983 \\
Deceased vs.\ Recov. & w/o diag & top-40 & SVC / LR & 0.880 & 0.882 & 0.880 & 0.880 & 0.938 \\
\addlinespace
Recov.\ vs.\ Amelior. & diag & top-20 & HistGB & 0.780 & 0.780 & 0.780 & 0.780 & 0.865 \\
Recov.\ vs.\ Amelior. & w/o diag & top-40 & LR & 0.690 & 0.691 & 0.690 & 0.690 & 0.730 \\
\bottomrule
\end{tabular}
\end{table}

Our complete results performed on all subsets, variants, and case studies are illustrated in the Supplementary Section (Supplementary Figures~S2--S4 and Supplementary Tables~S3--S5).

\subsection{Model Performance}

\begin{figure}[ht]
\centering
\includegraphics[width=0.95\textwidth]{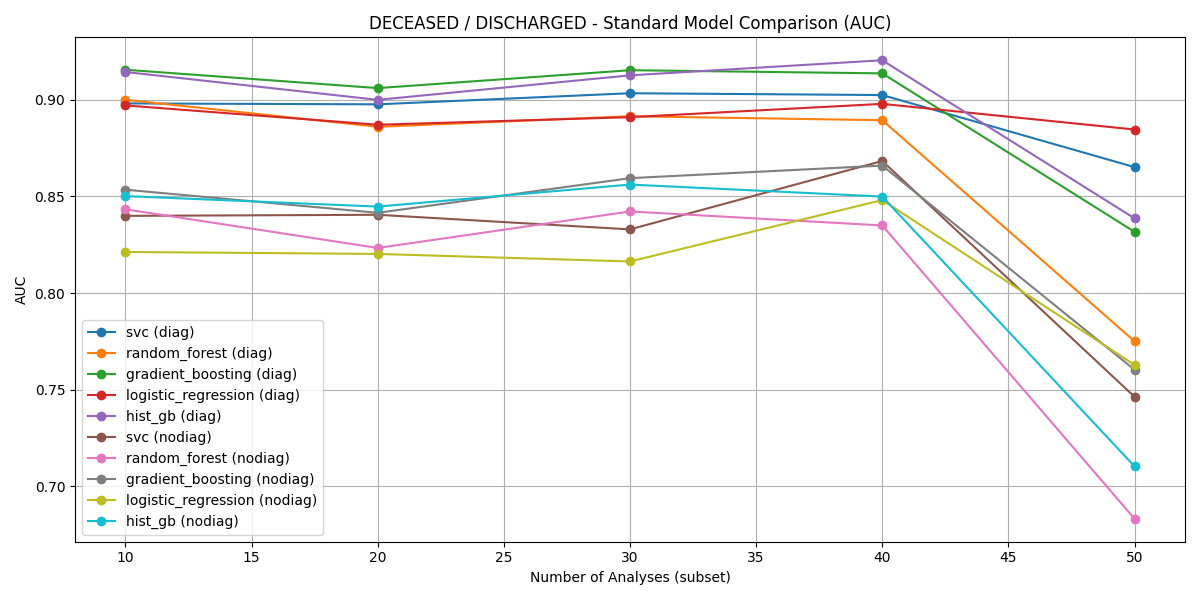}
\caption{AUC comparison of individual classifiers for Task~1---Deceased vs.\ Discharged. Diagnostic feature sets (``diag'') yield higher performance than non-diagnostic ones (``nodiag''), with Gradient Boosting variants achieving the best overall results.}
\label{fig:task1_auc}
\end{figure}

The most clinically significant case study focuses on predicting whether a patient will survive hospitalization or not. This binary classification between ``Deceased'' and ``Discharged'' (Alive) underscores the necessity of early prognosis in sepsis management, in contrast to the preceding two scenarios that examine more granular differences between distinct discharge types. For this task, the HistGB classifier peaked with an AUC of 0.9204 and accuracy of 0.843 on the top-40 diagnostic subset, representing the highest performance of any model in our study. The performance trends are visualized in Figure~\ref{fig:task1_auc}, which plots the AUC scores for all base models against the number of included laboratory tests.

\begin{figure}[ht]
\centering
\includegraphics[width=0.95\textwidth]{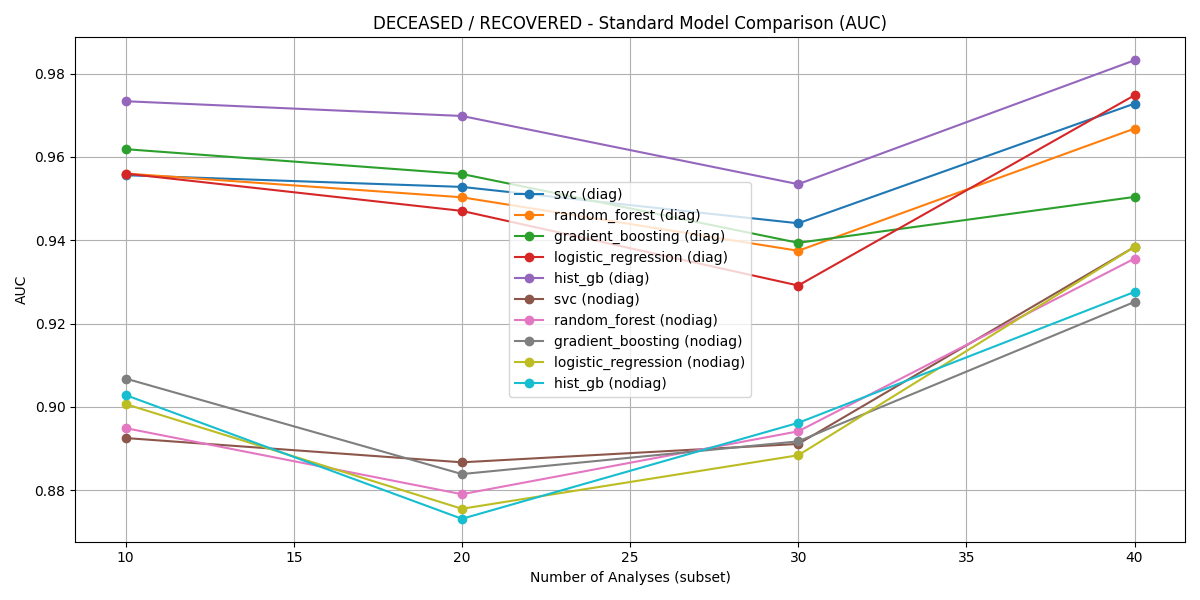}
\caption{Model performance comparison (AUC) for Deceased vs.\ Recovered patients across diagnostic subsets.}
\label{fig:task2_auc}
\end{figure}

The second case study, Deceased vs.\ Recovered, achieved the peak scores in accuracy and AUC. The top performance was achieved by a Histogram-based Gradient Boosting (HistGB) model using the top-30 investigations subset, scoring an AUC of 0.983 and accuracy and recall of 0.93 (Figure~\ref{fig:task2_auc}).

The second classification task investigated the distinction between patients who were discharged in a recovered state and those discharged in an ameliorated condition. While less critical than mortality prediction, this scenario helps assess whether machine learning models can identify patients likely to achieve full recovery versus those showing only partial improvement. From a clinical perspective, this task has value in setting realistic expectations for post-discharge outcomes and planning resource allocation.

\begin{figure}[ht]
\centering
\includegraphics[width=0.95\textwidth]{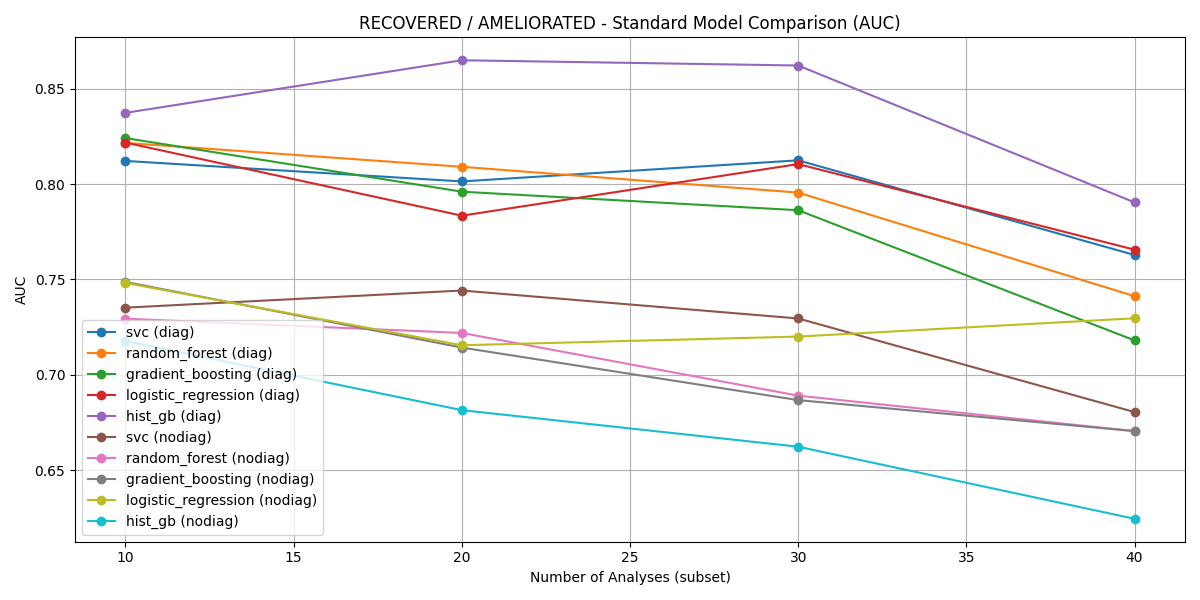}
\caption{Model performance comparison (AUC) for Recovered vs.\ Ameliorated patients across diagnostic subsets. Histogram-based Gradient Boosting models achieve the highest AUC values across all subsets.}
\label{fig:task3_auc}
\end{figure}

At last, the third case study, Recovered vs.\ Ameliorated, studies the finest details in optimistic outcomes. This scenario helps assess whether machine learning models can identify patients likely to achieve full recovery versus those showing only partial improvement. Here, the highest AUC was 0.865, achieved by Gradient Boosting on the top-20 diagnostic subset, while the Gradient Boosting classifier with diagnostics for top-30 lab tests achieved similar performance (Figure~\ref{fig:task3_auc}).

\subsection{SHAP Interpretation and Clinical Significance}

\begin{figure}[ht]
\centering
\includegraphics[width=0.68\textwidth]{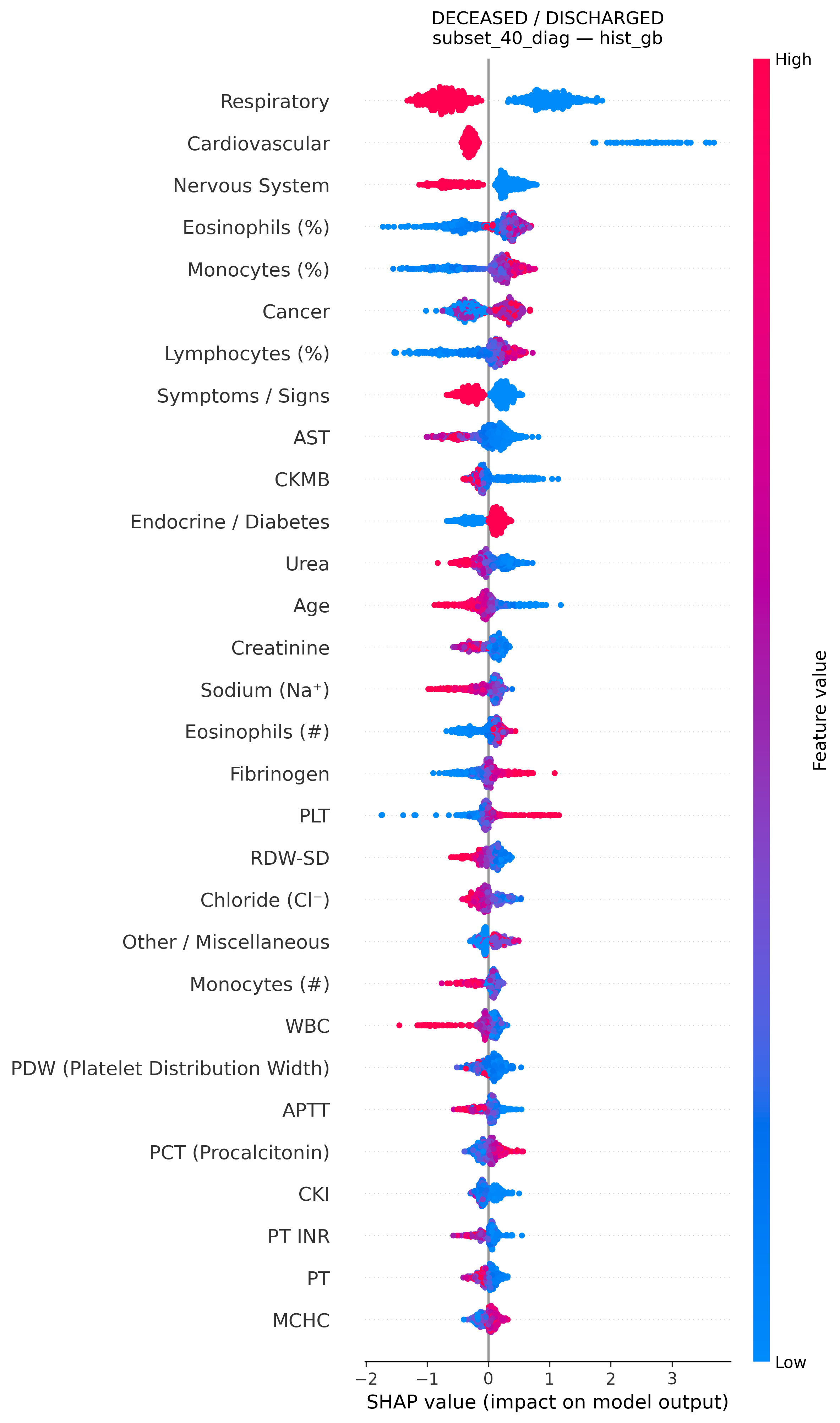}
\caption{SHAP summary plot for Deceased vs.\ Discharged top classifier (HistGB model trained on the top-40 diagnostic subset). Features are ranked by impact on mortality prediction.}
\label{fig:shap}
\end{figure}

We used SHAP (SHapley Additive exPlanations) to analyze the impact of the strongest features on the predictions made by the best model. Figure~\ref{fig:shap} illustrates the SHAP feature importance for the Deceased vs.\ Discharged outcome, highlighting the most influential variables driving mortality predictions. The SHAP analyses for the other classification tasks are thoroughly presented in the Supplementary Materials (Supplementary Figures~S5 and~S6).

All three SHAP analyses share a common set of important predictors in assessing the risk. Respiratory comorbidities are the top mortality threat, with Cardiovascular coming second, while patients with Digestive pathologies are more likely to make a full recovery. Patients with Nervous System comorbidities have an increased mortality risk, and overall younger patients are more likely to be discharged and make a full recovery.

Remarkably, our key finding is the undeniable importance of the Eosinophils across all tasks, and all variations of predictors. Decreased eosinophil counts, or eosinopenia, are becoming more widely acknowledged as an early, sensitive indicator of acute infection and systemic inflammation, even though they are not typically highlighted in sepsis scoring systems (SOFA, qSOFA). Our studies consolidate the supposition that lower proportions of Eosinophils among total white blood cells are inversely linked to fatal outcomes (lower the eosinophils percentage predict higher mortality risk).

Increased mortality risk was linked to elevated urea levels, which most likely reflected systemic catabolism and compromised kidney function. In line with older patients' weakened physiological reserves and immunological dysregulation, advancing age also made a substantial contribution. Poor outcomes are strongly predicted by sepsis-induced coagulopathy, which is indicated by low platelet counts. The models also recognize sodium levels as an important predictor. Both hyponatremia and hypernatremia are frequent in critically ill patients and can indicate underlying problems like kidney issues, dehydration, or hormonal imbalances.

The mortality-focused predictors placed greater weight on systemic inflammatory and hematologic signals (WBC, Eosinophils, Lymphocytes, and Monocytes; Fibrinogen; Platelet metrics including PLT/MPV/PDW) and on cardiac injury biomarkers (CKMB/CKI), whereas the recovery-oriented comparison (Recovered vs.\ Ameliorated) emphasizes hepatic and biliary function (ALT/AST and bilirubin fractions) and red-cell indices. Results indicate that liver stress or injury (elevated levels of AST---alanine aminotransferase) lead to a higher risk of mortality in this patient group. This aligns with widespread knowledge about sepsis, where liver dysfunction is a common and serious complication.

The SHAP analysis shows that the models rely on established markers of sepsis severity, such as liver enzymes and platelets, while also emphasizing the importance of less traditional but clinically plausible features like eosinophils, sodium levels, and specific comorbidity profiles. This suggests that the model has developed a detailed and clinically coherent approach to risk stratification. The SHAP analysis therefore reinforces the models' clinical applicability and transparency.

\section{Discussion}

As it was generally observed across several other studies~\cite{diwan2025,zhang2024}, gradient boosting models typically provide the best sepsis mortality prediction metrics compared to other supervised classification techniques.

Overall, this exhaustive model benchmarking underlines two important findings. First, machine learning models, particularly tree-based models, can offer exceptionally accurate mortality risk stratification (even better if the ICD-10 codes from diagnostics are available). Second, performance is competitive even in constrained settings, indicating usefulness in actual hospitals where lab completeness may differ.

Previous studies investigating ML-based sepsis outcome predictions~\cite{bao2023,zeng2021,hou2020} use the Medical Information Mart for Intensive Care (MIMIC III and IV) and the Intensive Care Unit Collaborative Research Database (eICU-CRD)---datasets from the United States.

However, the performance of models trained using this data may vary widely when applied on a population from a different geographical area. Factors that could influence such differences are yet unclear and require further exploration and may be related to country-specific healthcare environment, treatment choices, and standard of care. As we aimed to develop models that reflect as closely as possible the Romanian healthcare setting, we chose to train our models using a collection of patients with sepsis from the largest emergency hospital in Romania. To the authors' knowledge, this is the first study investigating sepsis outcome prediction using Machine Learning, based on a large collection of patients from an Eastern European country.

Patient status at discharge recorded in our dataset allowed us to explore two additional prediction scenarios, which have not been previously analyzed through ML techniques in patients with sepsis. In addition to mortality prediction, this analysis aimed to capture more subtle clinical transitions. These trends suggest that ML models are able to adapt beyond the binary ``dead or alive'' classification and are able to identify not only patients with sepsis who are likely to survive and fully recover but also patients with sepsis who could potentially improve. These types of models can help to inform early decisions of transfer from the ICU of patients who are stabilized and likely to continue improving, thus prioritizing intensive care resources in situations where these are very limited.

Prediction of outcomes reflecting patient deterioration or amelioration has been previously studied using ML techniques in a broader patient population, not restricted to sepsis~\cite{brankovic2022,steitz2023,yuan2025}. Akel and collaborators~\cite{akel2021} developed a tool for predicting clinical deterioration in adult hospitalized patients. Their GB model used only 24-hour trended heart rate, respiratory rate, and patient age to predict a composite outcome (probability of ICU transfer, death, or the combination of the two) and achieved an AUC of 0.79 to 0.80, significantly higher compared to conventional scores. In another study, Thiele and collaborators~\cite{thiele2025} trained ML models to predict patients in the ICU at risk of deterioration (defined as qSOFA$\geq$2), using vital signs information continuously recorded over a 24-hour period. Their model outperformed the RF, SVM, linear discriminant analysis, and LR models, with an AUC of 0.81 compared to 0.78, 0.78, 0.77, and 0.76, respectively.

In our study, the Deceased vs.\ Recovered classification achieved the highest overall performance, with the HistGradientBoosting model reaching an AUC of 0.983 and an accuracy and recall of 0.93. These results surpass previously reported performances on comparable sepsis mortality prediction tasks based on the MIMIC and eICU-CRD datasets (0.94--0.96 AUC~\cite{zhang2024,bao2023}). The Deceased vs.\ Discharged task performance also remained high (0.92 AUC), proving the robustness of our methodology. Moreover, the Recovered vs.\ Ameliorated classifier---a task not explored in sepsis works involving ML---managed to discern subtle recovery patterns (0.86 AUC), suggesting potential for early identification of patients likely to improve which can therefore be safely transferred from intensive care.

Routine laboratory results have been previously shown to play an important role in sepsis mortality prediction. This takes advantage of the ability of ML models to analyze large and complex sets of information, identifying relationships between patient characteristics unaccounted for by commonly used prognostic tools.

In accordance with other studies, SHAP analysis of previous studies revealed that the most important features for ML models are also some of those often used in risk scores, including urea, platelet count, and bilirubin. Moreover, most of the key variables highly relevant for outcome prediction in our scenarios were also highlighted in other studies, such as age~\cite{hu2022,he2024}, urea levels~\cite{zhang2024}, lymphocyte percentage~\cite{zhang2024lymph,lin2024}, platelet count, bilirubin~\cite{park2024,fan2024}, chloride levels~\cite{li2025}, and AST and ALT~\cite{choi2025}.

Even though the investigation showing the percentage of eosinophils among all white blood cells is only the 26th most frequently performed test in this study, it has proven to be one of the strongest predictors across all experiments. This aligns with previous evidence by Pinte \textit{et al.}\ (2025)~\cite{pinte2025}, who performed a systematic review and meta-analysis and found that patients hospitalized with infection who were presenting low eosinophil counts had a significantly higher mortality rate. Moreover, survivors showed an eosinophil count increase within the first 2--5 days of hospitalization, whereas the patients in critical condition persisted with low eosinophil counts.

Eosinopenia has been recognized in literature starting from the 19th century as part of the stress response to acute infections, even though it is not included in standardized sepsis assessment scores. The biological phenomenon---where inflammation and acute infection activate chemokines and stress hormones, leading to reduced blood eosinophil counts---has been repeatedly confirmed as an early indicator of immune dysregulation during sepsis hospitalization~\cite{abidi2008,shravani2025,alduhailib2021}.

This offers a clinical rationale for our results showing that patients with lowered eosinophils are likely to develop an intense inflammatory response, signaling higher risk of organ dysfunction and mortality. Recognizing eosinopenia as an early alarm signal could help the identification of septic patients in critical condition sooner, complementing other markers like platelets, hemoglobin, and lymphocytes, but at essentially no extra cost since eosinophils are part of routine blood counts.

The results presented should be interpreted taking into account some limitations. Only patients with complete data for each top-$N$ investigation subset were included in model development. The authors sought to balance variable count and sample size but did so using only five discrete scenarios (top-10, -20, -30, -40, and -50 most frequent laboratory tests). Moreover, the status at discharge for patients who were alive at discharge was subjectively recorded at the discretion of the treating physician. In future work, we plan to explore time-aware neural networks to better capture dynamic trends in sepsis progression following the findings of this work on feature impact.

\section{Conclusion}

The ML tools developed in this study demonstrated promising performance in predicting in-hospital mortality. Moreover, we demonstrated the potential to distinguish patients who are likely to achieve full recovery and those expected to show only partial post-sepsis improvement. Our findings add to emerging evidence supporting the value of ML-based approaches for clinical decision support, especially in resource-strained healthcare systems.

Given the scarcity of prior studies from Eastern Europe, this study contributes to understanding regional sepsis characteristics. Integrating our models into clinical workflows could complement existing scores such as SOFA in assisting clinicians in triage, thus ultimately improving survival through earlier intervention, while showing transparent predictions based on familiar variables---such as enzymes and eosinophils, and diagnostic or urea levels.

\section*{Author Contributions}
The first two authors, Bunea, A.A.\ and Mu\c{s}at, F., had equal contributions to this work. Mu\c{s}at, F.\ oversaw dataset collection and preparation and provided the relevant medical expertise. Bunea, A.A.\ was responsible for dataset preprocessing, model training, evaluation, and comparative analysis of the machine learning models. The discussion, interpretation of results, and manuscript preparation were carried out collaboratively by all co-authors.

\section*{Data Availability}
The datasets included in this study are not publicly available due to patient privacy regulations and ethical restrictions but are available from the corresponding author on reasonable request.

\section*{Competing Interests}
The authors declare no competing interests.

\section*{Funding}
This research received no external funding at the time of submission. An acknowledgement of funding support will be added once the grant identification details are formally assigned.


\clearpage
\appendix
\setcounter{table}{0}
\setcounter{figure}{0}
\renewcommand{\thetable}{S\arabic{table}}
\renewcommand{\thefigure}{S\arabic{figure}}

\section*{Supplementary Materials}

\subsection*{S1. Dataset and Laboratory Test Coverage}

Each diagnostic was associated with a high-level comorbidity category by its leading alphabetical character from the ICD-10 convention. Following this rationale, we categorized the diagnostics into 14 main comorbidity categories. For example, ICD codes starting with A or B have been classified as Infectious Diseases (e.g., A41.9---Sepsis, B20---HIV), C and D as Cancer (e.g., C18.9---Colon Cancer, D45---Myeloproliferative Neoplasm), E as Endocrine Disorders (e.g., Type 2 Diabetes), and so on. The mapping used to convert ICD-10 diagnosis codes into clinically relevant comorbidity categories is summarized in Table~\ref{tab:supp_icd}.

\begin{table}[ht]
\centering
\caption{Mapping of ICD-10 diagnostic code initials to comorbidity categories.}
\label{tab:supp_icd}
\small
\begin{tabular}{lll}
\toprule
\textbf{ICD-10 Prefix} & \textbf{Comorbidity Category} & \textbf{Examples} \\
\midrule
A, B & Infectious Diseases & A41.9 Sepsis, B20 HIV \\
C, D & Cancer & C18.9 Colon Cancer, D45 Myeloproliferative Neoplasm \\
E & Endocrine / Diabetes & E11 Type 2 Diabetes \\
F & Mental Health & F32 Depressive Episode, F20 Schizophrenia \\
G & Nervous System & G40 Epilepsy, G20 Parkinson's Disease \\
H & Eye / Ear Disorders & H25 Cataract, H66 Otitis Media \\
I & Cardiovascular & I21 Myocardial Infarction, I50 Heart Failure \\
J & Respiratory & J18.0 Pneumonia, J44 COPD \\
K & Digestive System & K74 Liver Cirrhosis, K35 Appendicitis \\
L & Skin Disorders & L03 Cellulitis, L40 Psoriasis \\
M & Musculoskeletal & M17 Knee Osteoarthritis, M79.7 Fibromyalgia \\
N & Renal / Urinary & N18 Chronic Kidney Disease, N10 Acute Pyelonephritis \\
O & Pregnancy / Obstetric & O80 Normal Delivery, O14 Preeclampsia \\
R & Symptoms / Signs & R50 Fever, R55 Syncope \\
T--Z & Other / Miscellaneous & T81 Postoperative Complications, Z51 Chemotherapy \\
\bottomrule
\end{tabular}
\end{table}

\begin{longtable}{clrr}
\caption{Distribution of patients for top 50 most frequent laboratory tests.} \label{tab:supp_labs} \\
\toprule
\textbf{\#} & \textbf{Laboratory Test} & \textbf{Count} & \textbf{\% of patients} \\
\midrule
\endfirsthead
\caption[]{(continued)} \\
\toprule
\textbf{\#} & \textbf{Laboratory Test} & \textbf{Count} & \textbf{\% of patients} \\
\midrule
\endhead
\bottomrule
\endfoot
1 & Urea & 12,216 & 99.33\% \\
2 & Sodium (Na$^+$) & 12,214 & 99.32\% \\
3 & Potassium (K$^+$) & 12,212 & 99.30\% \\
4 & Glucose & 12,209 & 99.28\% \\
5 & AST (Aspartate Aminotransferase) & 12,193 & 99.15\% \\
6 & ALT (Alanine Aminotransferase) & 12,192 & 99.14\% \\
7 & PT INR (International Normalized Ratio) & 11,905 & 96.80\% \\
8 & Total Bilirubin & 11,687 & 95.03\% \\
9 & PT (Prothrombin Time, sec) & 11,683 & 95.00\% \\
10 & APTT (Activated Partial Thromboplastin Time, sec) & 11,678 & 94.96\% \\
11 & PT (\%) (Prothrombin Time, \%) & 11,662 & 94.83\% \\
12 & Direct Bilirubin & 11,229 & 91.31\% \\
13 & CKMB (Creatine Kinase-MB) & 11,110 & 90.34\% \\
14 & Amylase & 10,975 & 89.24\% \\
15 & Chloride (Cl$^-$) & 10,863 & 88.33\% \\
16 & Hemogram MCV (Mean Corpuscular Volume) & 10,806 & 87.87\% \\
17 & Hemogram WBC (White Blood Cells) & 10,806 & 87.87\% \\
18 & Hemogram RBC (Red Blood Cells) & 10,806 & 87.87\% \\
19 & Hemogram MCH (Mean Corpuscular Hemoglobin) & 10,806 & 87.87\% \\
20 & Hemogram MPV (Mean Platelet Volume) & 10,768 & 87.56\% \\
21 & Hemogram HGB (Hemoglobin) & 10,765 & 87.53\% \\
22 & Hemogram PLT (Platelets) & 10,765 & 87.53\% \\
23 & Hemogram HCT (Hematocrit) & 10,765 & 87.53\% \\
24 & Hemogram RDW (Red Cell Distribution Width) & 10,500 & 85.38\% \\
25 & Hemogram MCHC (Mean Corpusc.\ Hemoglobin Conc.) & 10,309 & 83.83\% \\
26 & Hemogram Eosinophils (\%) & 9,903 & 80.53\% \\
27 & Hemogram Eosinophils (\#) & 9,752 & 79.30\% \\
28 & Creatinine & 9,389 & 76.35\% \\
29 & CKI (Creatine Kinase Index) & 9,345 & 75.99\% \\
30 & Hemogram RDW-SD (Red Cell Distribution Width -- SD) & 8,916 & 72.50\% \\
31 & Lipase & 8,641 & 70.26\% \\
32 & Hemogram PDW (Platelet Distribution Width) & 8,294 & 67.44\% \\
33 & Hemogram PCT (Procalcitonin) & 8,144 & 66.22\% \\
34 & Indirect Bilirubin & 8,117 & 66.00\% \\
35 & Fibrinogen & 8,030 & 65.30\% \\
36 & Hemogram Lymphocytes (\%) & 7,498 & 60.97\% \\
37 & Hemogram Monocytes (\%) & 7,498 & 60.97\% \\
38 & Hemogram Basophils (\#) & 7,397 & 60.15\% \\
39 & Hemogram Monocytes (\#) & 7,397 & 60.15\% \\
40 & Hemogram Neutrophils (\#) & 7,397 & 60.15\% \\
41 & Hemogram Basophils (\%) & 7,397 & 60.15\% \\
42 & Hemogram Neutrophils (\%) & 7,397 & 60.15\% \\
43 & Hemogram Lymphocytes (\#) & 7,397 & 60.15\% \\
44 & Hemogram NRBC (\%) (Nucleated RBC \%) & 7,086 & 57.62\% \\
45 & Hemogram NRBC (\#) (Nucleated RBC \#) & 7,073 & 57.51\% \\
46 & Fibrinogen C & 6,051 & 49.20\% \\
47 & GGT (Gamma-Glutamyl Transferase) & 6,003 & 48.81\% \\
48 & Calcium (Ca) & 5,945 & 48.34\% \\
49 & Total Protein (TP) & 5,627 & 45.76\% \\
50 & CRP (C-Reactive Protein) & 5,357 & 43.56\% \\
\end{longtable}

\subsection*{S2. Machine Learning Models -- Technical Details}

\subsubsection*{Model Selection Rationale}
Our dataset consists of sparse complex data collecting demographic and discharge information, comorbidity screening and investigation results. For all the training setups, we create a standardized input and output format that is flexible to the number of input parameters of each trial. Running multiple experiments for different feature subsets, we ensure a scalable and robust training mechanism for inputs ranging from 14 to 70 parameters to predict a binary classification output. To ensure interpretable analysis of the feature weights, we selected widely adopted classical machine learning algorithms: Logistic Regression (LR), Support Vector Classifier (SVC), Random Forest (RF), Gradient Boosting (GB), and Histogram-based Gradient Boosting (HistGB). These models provide complementary approaches in solving the classification tasks, varying in methods from linear separators to non-linear tree ensembles.

\subsubsection*{Feature Engineering and Scaling}
The input for our model has mixed data types: continuous test results, label-encoded variables for sex and environment of origin, and optional structured clinical context found in comorbidities. Depending on the model used to perform the classification, some of these features were processed into equivalent distributions.

Continuous features (e.g., laboratory test results, age) were used in their raw form for tree-based models such as Random Forest and Gradient Boosting, which are insensitive to feature scaling and capable of capturing non-linear thresholds. For models sensitive to feature magnitudes, such as Logistic Regression and Support Vector Classifier, all continuous inputs were $z$-normalized to zero mean and unit variance to stabilize gradient-based optimization.

Through this preprocessing pipeline, each model received a standardized, fully numerical input vector tailored to its architectural requirements, while preserving clinical interpretability across experiments.

To investigate the impact of feature dimensionality in relation to the dataset size, we varied the number of investigations used to train the models. To find the optimal collection of laboratory tests, we fitted each model in subsets of the top 10, 20, 30, 40, and 50 lab tests ranked by their frequency in the hospitalization record as presented in Supplementary Table~\ref{tab:supp_labs}. Smaller subsets of investigations include only the most frequent laboratory tests, allowing the models to train on a larger patient cohort. In contrast, broader subsets provide additional insight but reduce the dataset, as we only include patients who underwent all the tests within the subset.

\subsubsection*{Training Protocol}
Models were trained on five sets of laboratory tests: the Top-10, 20, 30, 40, and 50 most frequent investigations. Each of the five subsets was trained on two input configurations: (1) \textbf{Base variant:} Demographics + lab results; (2) \textbf{Extended variant:} Demographics + lab results + comorbidities.

This design allowed us to analyze the impact of diagnosis-related features for all classification conditions.

To address class imbalance (e.g., higher survivors compared to deaths), we under-sampled the dominant class for each training divide. This maintains class balance and prevents models from converging on majority-bias solutions.

For each experiment environment (subset $\times$ variant $\times$ task), the data was separated on stratified 80/20 division for test and training sets. This ensured class distribution in both the sets and allowed for balanced analysis. A fixed random state seed was used for reproducibility.

All models were trained using a consistent pipeline architecture consisting of: a preprocessing layer (continuous features standardized using StandardScaler; categorical features one-hot encoded using OneHotEncoder) and a classifier layer (one of: Logistic Regression, Support Vector Classifier, Random Forest, Gradient Boosting).

The classifiers were evaluated individually and in ensemble using the same metrics: accuracy, F1-score, ROC AUC, and confusion matrix. The full training pipeline including preprocessing and trained classifier was saved as a \texttt{.joblib} object for each configuration, enabling seamless reuse and deployment.

\subsubsection*{Explainability Analysis}
One of the ultimate goals of this work was not simply to construct good classifiers but to interpret and justify their decisions in a clinically relevant manner as well. For each of the models, we computed feature-importance profiles to find which of the input variables was most responsible for the predictions made by the model.

For Gradient Boosting and Random Forest models, we derived the feature weights directly from the trained classifiers through the \texttt{feature\_importances\_} property of Scikit-learn. These scores represent the cumulative reduction in Gini impurity across all trees where a feature was used to split the data.

With Logistic Regression, we employed the absolute magnitude of the standardized coefficients as an influence proxy for each feature. As this model has input $z$-normalized, the resulting coefficients are equivalent on an absolute scale and describe how strongly each feature contributes to the predicted log-odds of the end-state.

The Support Vector Classifier model, with no inherent feature-importance outputs, was evaluated by the application of permutation importance.

\subsection*{S3. Complete Results Tables}

\begin{longtable}{cclccccc}
\caption{Performance of all base models for Task~1 (Deceased vs.\ Discharged).} \label{tab:supp_task1} \\
\toprule
\textbf{Subset} & \textbf{Variant} & \textbf{Model} & \textbf{Accuracy} & \textbf{Precision} & \textbf{Recall} & \textbf{F1} & \textbf{AUC} \\
\midrule
\endfirsthead
\caption[]{(continued)} \\
\toprule
\textbf{Subset} & \textbf{Variant} & \textbf{Model} & \textbf{Accuracy} & \textbf{Precision} & \textbf{Recall} & \textbf{F1} & \textbf{AUC} \\
\midrule
\endhead
\bottomrule
\endfoot
10 & diag & SVC & 0.8184 & 0.8184 & 0.8184 & 0.8184 & 0.8981 \\
10 & diag & Random Forest & 0.8178 & 0.8178 & 0.8178 & 0.8178 & 0.9000 \\
10 & diag & Gradient Boosting & 0.8339 & 0.8339 & 0.8339 & 0.8339 & 0.9155 \\
10 & diag & Logistic Regression & 0.8051 & 0.8052 & 0.8051 & 0.8051 & 0.8971 \\
10 & diag & HistGB & 0.8350 & 0.8351 & 0.8350 & 0.8350 & 0.9144 \\
10 & nodiag & SVC & 0.7658 & 0.7667 & 0.7658 & 0.7656 & 0.8399 \\
10 & nodiag & Random Forest & 0.7658 & 0.7667 & 0.7658 & 0.7656 & 0.8433 \\
10 & nodiag & Gradient Boosting & 0.7702 & 0.7705 & 0.7702 & 0.7701 & 0.8535 \\
10 & nodiag & Logistic Regression & 0.7403 & 0.7425 & 0.7403 & 0.7397 & 0.8213 \\
10 & nodiag & HistGB & 0.7719 & 0.7720 & 0.7719 & 0.7718 & 0.8502 \\
\addlinespace
20 & diag & SVC & 0.8127 & 0.8127 & 0.8127 & 0.8127 & 0.8976 \\
20 & diag & Random Forest & 0.8079 & 0.8082 & 0.8079 & 0.8079 & 0.8860 \\
20 & diag & Gradient Boosting & 0.8208 & 0.8208 & 0.8208 & 0.8208 & 0.9060 \\
20 & diag & Logistic Regression & 0.8018 & 0.8018 & 0.8018 & 0.8018 & 0.8871 \\
20 & diag & HistGB & 0.8195 & 0.8195 & 0.8195 & 0.8195 & 0.9000 \\
20 & nodiag & SVC & 0.7616 & 0.7619 & 0.7616 & 0.7615 & 0.8405 \\
20 & nodiag & Random Forest & 0.7527 & 0.7530 & 0.7527 & 0.7527 & 0.8233 \\
20 & nodiag & Gradient Boosting & 0.7616 & 0.7617 & 0.7616 & 0.7616 & 0.8415 \\
20 & nodiag & Logistic Regression & 0.7296 & 0.7301 & 0.7296 & 0.7294 & 0.8202 \\
20 & nodiag & HistGB & 0.7582 & 0.7586 & 0.7582 & 0.7581 & 0.8447 \\
\addlinespace
30 & diag & SVC & 0.8270 & 0.8271 & 0.8270 & 0.8270 & 0.9034 \\
30 & diag & Random Forest & 0.8101 & 0.8102 & 0.8101 & 0.8101 & 0.8915 \\
30 & diag & Gradient Boosting & 0.8330 & 0.8330 & 0.8330 & 0.8330 & 0.9153 \\
30 & diag & Logistic Regression & 0.8161 & 0.8163 & 0.8161 & 0.8161 & 0.8910 \\
30 & diag & HistGB & 0.8380 & 0.8381 & 0.8380 & 0.8380 & 0.9126 \\
30 & nodiag & SVC & 0.7694 & 0.7694 & 0.7694 & 0.7694 & 0.8329 \\
30 & nodiag & Random Forest & 0.7535 & 0.7535 & 0.7535 & 0.7535 & 0.8422 \\
30 & nodiag & Gradient Boosting & 0.7694 & 0.7697 & 0.7694 & 0.7693 & 0.8594 \\
30 & nodiag & Logistic Regression & 0.7336 & 0.7336 & 0.7336 & 0.7336 & 0.8163 \\
30 & nodiag & HistGB & 0.7694 & 0.7694 & 0.7694 & 0.7694 & 0.8561 \\
\addlinespace
40 & diag & SVC & 0.8294 & 0.8294 & 0.8294 & 0.8294 & 0.9024 \\
40 & diag & Random Forest & 0.8106 & 0.8110 & 0.8106 & 0.8105 & 0.8894 \\
40 & diag & Gradient Boosting & 0.8208 & 0.8209 & 0.8208 & 0.8208 & 0.9137 \\
40 & diag & Logistic Regression & 0.8225 & 0.8225 & 0.8225 & 0.8225 & 0.8979 \\
40 & diag & HistGB & \textbf{0.8430} & \textbf{0.8430} & \textbf{0.8430} & \textbf{0.8430} & \textbf{0.9204} \\
40 & nodiag & SVC & 0.7799 & 0.7803 & 0.7799 & 0.7798 & 0.8683 \\
40 & nodiag & Random Forest & 0.7662 & 0.7667 & 0.7662 & 0.7661 & 0.8350 \\
40 & nodiag & Gradient Boosting & 0.7935 & 0.7935 & 0.7935 & 0.7935 & 0.8659 \\
40 & nodiag & Logistic Regression & 0.7713 & 0.7713 & 0.7713 & 0.7713 & 0.8481 \\
40 & nodiag & HistGB & 0.7782 & 0.7784 & 0.7782 & 0.7781 & 0.8499 \\
\addlinespace
50 & diag & SVC & 0.7569 & 0.7571 & 0.7569 & 0.7568 & 0.8652 \\
50 & diag & Random Forest & 0.6972 & 0.6978 & 0.6972 & 0.6970 & 0.7751 \\
50 & diag & Gradient Boosting & 0.7431 & 0.7432 & 0.7431 & 0.7431 & 0.8317 \\
50 & diag & Logistic Regression & 0.7936 & 0.7936 & 0.7936 & 0.7936 & 0.8846 \\
50 & diag & HistGB & 0.7477 & 0.7487 & 0.7477 & 0.7474 & 0.8386 \\
50 & nodiag & SVC & 0.6514 & 0.6516 & 0.6514 & 0.6513 & 0.7463 \\
50 & nodiag & Random Forest & 0.6468 & 0.6474 & 0.6468 & 0.6464 & 0.6832 \\
50 & nodiag & Gradient Boosting & 0.7018 & 0.7027 & 0.7018 & 0.7015 & 0.7603 \\
50 & nodiag & Logistic Regression & 0.6972 & 0.6978 & 0.6972 & 0.6970 & 0.7628 \\
50 & nodiag & HistGB & 0.6606 & 0.6606 & 0.6606 & 0.6605 & 0.7103 \\
\end{longtable}

\begin{longtable}{cclccccc}
\caption{Performance of all base models for Task~2 (Deceased vs.\ Recovered).} \label{tab:supp_task2} \\
\toprule
\textbf{Subset} & \textbf{Variant} & \textbf{Model} & \textbf{Accuracy} & \textbf{Precision} & \textbf{Recall} & \textbf{F1} & \textbf{AUC} \\
\midrule
\endfirsthead
\caption[]{(continued)} \\
\toprule
\textbf{Subset} & \textbf{Variant} & \textbf{Model} & \textbf{Accuracy} & \textbf{Precision} & \textbf{Recall} & \textbf{F1} & \textbf{AUC} \\
\midrule
\endhead
\bottomrule
\endfoot
10 & diag & SVC & 0.8767 & 0.8768 & 0.8767 & 0.8767 & 0.9556 \\
10 & diag & Random Forest & 0.8814 & 0.8824 & 0.8814 & 0.8813 & 0.9560 \\
10 & diag & Gradient Boosting & 0.8930 & 0.8931 & 0.8930 & 0.8930 & 0.9619 \\
10 & diag & Logistic Regression & 0.8674 & 0.8676 & 0.8674 & 0.8674 & 0.9560 \\
10 & diag & HistGB & 0.9116 & 0.9118 & 0.9116 & 0.9116 & 0.9734 \\
10 & nodiag & SVC & 0.8140 & 0.8140 & 0.8140 & 0.8140 & 0.8925 \\
10 & nodiag & Random Forest & 0.8023 & 0.8034 & 0.8023 & 0.8021 & 0.8949 \\
10 & nodiag & Gradient Boosting & 0.8047 & 0.8047 & 0.8047 & 0.8046 & 0.9068 \\
10 & nodiag & Logistic Regression & 0.8116 & 0.8117 & 0.8116 & 0.8116 & 0.9007 \\
10 & nodiag & HistGB & 0.7953 & 0.7956 & 0.7953 & 0.7953 & 0.9028 \\
\addlinespace
20 & diag & SVC & 0.8882 & 0.8882 & 0.8882 & 0.8882 & 0.9528 \\
20 & diag & Random Forest & 0.8722 & 0.8724 & 0.8722 & 0.8722 & 0.9503 \\
20 & diag & Gradient Boosting & 0.8914 & 0.8920 & 0.8914 & 0.8913 & 0.9559 \\
20 & diag & Logistic Regression & 0.8658 & 0.8664 & 0.8658 & 0.8658 & 0.9470 \\
20 & diag & HistGB & 0.8914 & 0.8920 & 0.8914 & 0.8913 & 0.9698 \\
20 & nodiag & SVC & 0.8083 & 0.8083 & 0.8083 & 0.8083 & 0.8867 \\
20 & nodiag & Random Forest & 0.7987 & 0.7997 & 0.7987 & 0.7985 & 0.8790 \\
20 & nodiag & Gradient Boosting & 0.8083 & 0.8087 & 0.8083 & 0.8082 & 0.8839 \\
20 & nodiag & Logistic Regression & 0.7923 & 0.7924 & 0.7923 & 0.7923 & 0.8756 \\
20 & nodiag & HistGB & 0.8051 & 0.8051 & 0.8051 & 0.8051 & 0.8731 \\
\addlinespace
30 & diag & SVC & 0.8500 & 0.8521 & 0.8500 & 0.8498 & 0.9441 \\
30 & diag & Random Forest & 0.8500 & 0.8553 & 0.8500 & 0.8494 & 0.9375 \\
30 & diag & Gradient Boosting & 0.8667 & 0.8674 & 0.8667 & 0.8666 & 0.9394 \\
30 & diag & Logistic Regression & 0.8278 & 0.8311 & 0.8278 & 0.8273 & 0.9291 \\
30 & diag & HistGB & 0.8833 & 0.8857 & 0.8833 & 0.8832 & 0.9535 \\
30 & nodiag & SVC & 0.7889 & 0.7912 & 0.7889 & 0.7885 & 0.8911 \\
30 & nodiag & Random Forest & 0.8167 & 0.8199 & 0.8167 & 0.8162 & 0.8941 \\
30 & nodiag & Gradient Boosting & 0.7944 & 0.7974 & 0.7944 & 0.7939 & 0.8917 \\
30 & nodiag & Logistic Regression & 0.8056 & 0.8065 & 0.8056 & 0.8054 & 0.8884 \\
30 & nodiag & HistGB & 0.7944 & 0.7962 & 0.7944 & 0.7941 & 0.8962 \\
\addlinespace
40 & diag & SVC & 0.8900 & 0.8939 & 0.8900 & 0.8897 & 0.9728 \\
40 & diag & Random Forest & 0.9100 & 0.9102 & 0.9100 & 0.9100 & 0.9668 \\
40 & diag & Gradient Boosting & 0.8500 & 0.8513 & 0.8500 & 0.8499 & 0.9504 \\
40 & diag & Logistic Regression & 0.8900 & 0.8902 & 0.8900 & 0.8900 & 0.9748 \\
40 & diag & HistGB & \textbf{0.9300} & \textbf{0.9316} & \textbf{0.9300} & \textbf{0.9299} & \textbf{0.9832} \\
40 & nodiag & SVC & 0.8800 & 0.8806 & 0.8800 & 0.8800 & 0.9384 \\
40 & nodiag & Random Forest & 0.8300 & 0.8301 & 0.8300 & 0.8300 & 0.9356 \\
40 & nodiag & Gradient Boosting & 0.8300 & 0.8301 & 0.8300 & 0.8300 & 0.9252 \\
40 & nodiag & Logistic Regression & 0.8800 & 0.8824 & 0.8800 & 0.8798 & 0.9384 \\
40 & nodiag & HistGB & 0.8200 & 0.8200 & 0.8200 & 0.8200 & 0.9276 \\
\end{longtable}

\begin{longtable}{cclccccc}
\caption{Performance of all base models for Task~3 (Recovered vs.\ Ameliorated).} \label{tab:supp_task3} \\
\toprule
\textbf{Subset} & \textbf{Variant} & \textbf{Model} & \textbf{Accuracy} & \textbf{Precision} & \textbf{Recall} & \textbf{F1} & \textbf{AUC} \\
\midrule
\endfirsthead
\caption[]{(continued)} \\
\toprule
\textbf{Subset} & \textbf{Variant} & \textbf{Model} & \textbf{Accuracy} & \textbf{Precision} & \textbf{Recall} & \textbf{F1} & \textbf{AUC} \\
\midrule
\endhead
\bottomrule
\endfoot
10 & diag & SVC & 0.7535 & 0.7626 & 0.7535 & 0.7513 & 0.8122 \\
10 & diag & Random Forest & 0.7465 & 0.7517 & 0.7465 & 0.7452 & 0.8216 \\
10 & diag & Gradient Boosting & 0.7442 & 0.7521 & 0.7442 & 0.7422 & 0.8241 \\
10 & diag & Logistic Regression & 0.7581 & 0.7665 & 0.7581 & 0.7562 & 0.8218 \\
10 & diag & HistGB & 0.7651 & 0.7707 & 0.7651 & 0.7639 & 0.8374 \\
10 & nodiag & SVC & 0.6628 & 0.6654 & 0.6628 & 0.6615 & 0.7352 \\
10 & nodiag & Random Forest & 0.6744 & 0.6750 & 0.6744 & 0.6742 & 0.7294 \\
10 & nodiag & Gradient Boosting & 0.6860 & 0.6875 & 0.6860 & 0.6854 & 0.7487 \\
10 & nodiag & Logistic Regression & 0.6860 & 0.6867 & 0.6860 & 0.6858 & 0.7481 \\
10 & nodiag & HistGB & 0.6605 & 0.6606 & 0.6605 & 0.6604 & 0.7177 \\
\addlinespace
20 & diag & SVC & 0.7348 & 0.7359 & 0.7348 & 0.7345 & 0.8014 \\
20 & diag & Random Forest & 0.7540 & 0.7557 & 0.7540 & 0.7535 & 0.8091 \\
20 & diag & Gradient Boosting & 0.7316 & 0.7318 & 0.7316 & 0.7316 & 0.7960 \\
20 & diag & Logistic Regression & 0.7220 & 0.7221 & 0.7220 & 0.7220 & 0.7834 \\
20 & diag & HistGB & \textbf{0.7796} & \textbf{0.7798} & \textbf{0.7796} & \textbf{0.7795} & \textbf{0.8650} \\
20 & nodiag & SVC & 0.6805 & 0.6806 & 0.6805 & 0.6804 & 0.7442 \\
20 & nodiag & Random Forest & 0.6773 & 0.6773 & 0.6773 & 0.6773 & 0.7219 \\
20 & nodiag & Gradient Boosting & 0.6454 & 0.6454 & 0.6454 & 0.6453 & 0.7142 \\
20 & nodiag & Logistic Regression & 0.6741 & 0.6742 & 0.6741 & 0.6741 & 0.7155 \\
20 & nodiag & HistGB & 0.6358 & 0.6358 & 0.6358 & 0.6358 & 0.6814 \\
\addlinespace
30 & diag & SVC & 0.7389 & 0.7396 & 0.7389 & 0.7387 & 0.8125 \\
30 & diag & Random Forest & 0.7389 & 0.7413 & 0.7389 & 0.7382 & 0.7956 \\
30 & diag & Gradient Boosting & 0.7278 & 0.7312 & 0.7278 & 0.7268 & 0.7863 \\
30 & diag & Logistic Regression & 0.7333 & 0.7363 & 0.7333 & 0.7325 & 0.8105 \\
30 & diag & HistGB & 0.7667 & 0.7700 & 0.7667 & 0.7659 & 0.8622 \\
30 & nodiag & SVC & 0.6556 & 0.6584 & 0.6556 & 0.6540 & 0.7295 \\
30 & nodiag & Random Forest & 0.6500 & 0.6523 & 0.6500 & 0.6487 & 0.6890 \\
30 & nodiag & Gradient Boosting & 0.6333 & 0.6377 & 0.6333 & 0.6304 & 0.6867 \\
30 & nodiag & Logistic Regression & 0.6722 & 0.6733 & 0.6722 & 0.6717 & 0.7200 \\
30 & nodiag & HistGB & 0.6278 & 0.6297 & 0.6278 & 0.6264 & 0.6622 \\
\addlinespace
40 & diag & SVC & 0.6900 & 0.6938 & 0.6900 & 0.6885 & 0.7628 \\
40 & diag & Random Forest & 0.7100 & 0.7170 & 0.7100 & 0.7076 & 0.7412 \\
40 & diag & Gradient Boosting & 0.6800 & 0.6847 & 0.6800 & 0.6779 & 0.7180 \\
40 & diag & Logistic Regression & 0.7300 & 0.7377 & 0.7300 & 0.7278 & 0.7656 \\
40 & diag & HistGB & 0.7600 & 0.7708 & 0.7600 & 0.7576 & 0.7904 \\
40 & nodiag & SVC & 0.6300 & 0.6305 & 0.6300 & 0.6297 & 0.6804 \\
40 & nodiag & Random Forest & 0.6300 & 0.6301 & 0.6300 & 0.6300 & 0.6704 \\
40 & nodiag & Gradient Boosting & 0.6600 & 0.6623 & 0.6600 & 0.6588 & 0.6704 \\
40 & nodiag & Logistic Regression & 0.6900 & 0.6907 & 0.6900 & 0.6897 & 0.7296 \\
40 & nodiag & HistGB & 0.6400 & 0.6437 & 0.6400 & 0.6377 & 0.6244 \\
\end{longtable}

\clearpage

\begin{figure}[ht]
\centering
\includegraphics[width=0.7\textwidth]{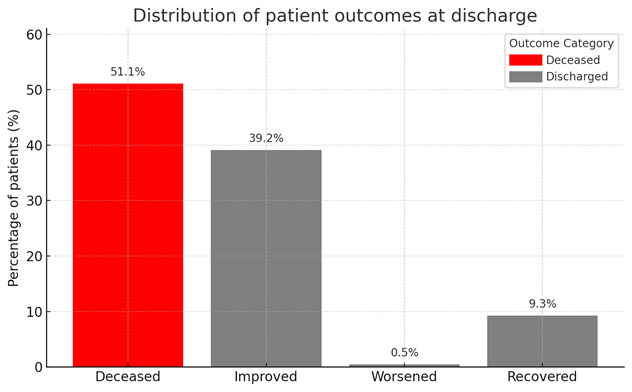}
\caption{Distribution of patient outcomes at discharge. Red indicates deceased patients, while gray represents discharged patients, including improved, recovered, and worsened cases.}
\label{fig:supp_outcomes}
\end{figure}

\begin{figure}[ht]
\centering
\includegraphics[width=0.95\textwidth]{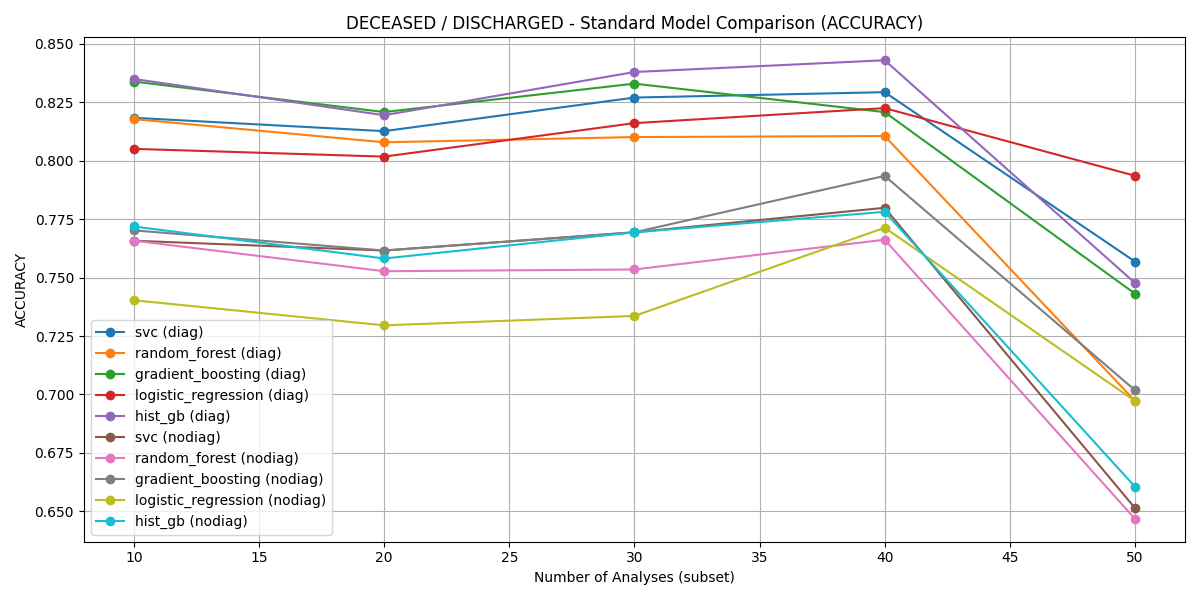}
\caption{Model accuracy comparison for Deceased vs.\ Discharged patients across diagnostic subsets.}
\label{fig:supp_task1_acc}
\end{figure}

\begin{figure}[ht]
\centering
\includegraphics[width=0.95\textwidth]{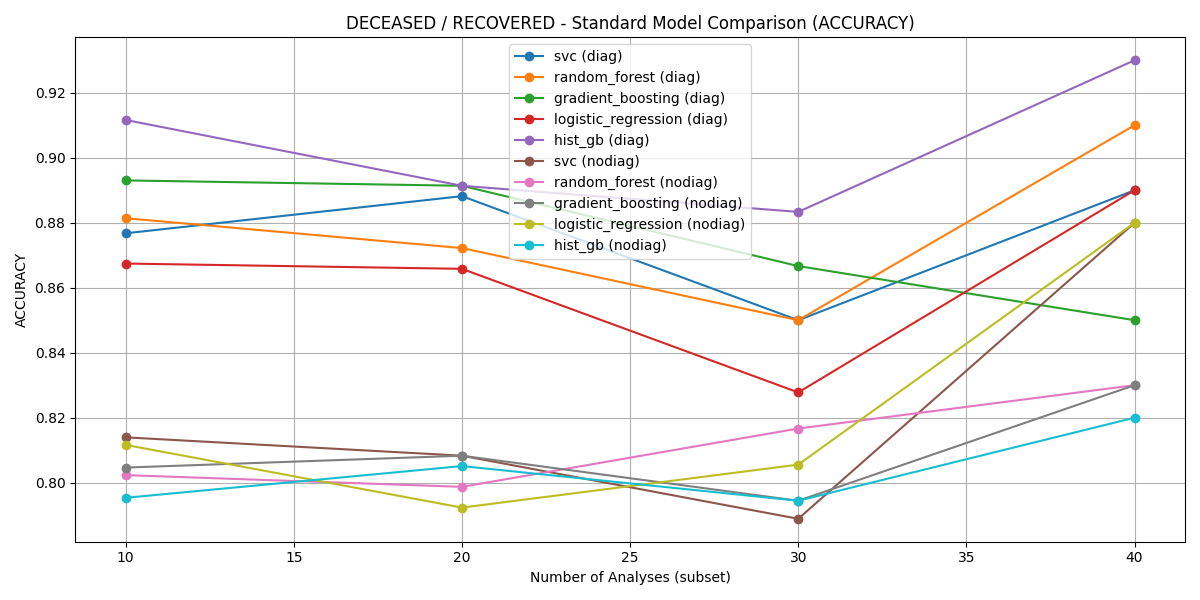}
\caption{Model accuracy comparison for Deceased vs.\ Recovered patients across diagnostic subsets.}
\label{fig:supp_task2_acc}
\end{figure}

\begin{figure}[ht]
\centering
\includegraphics[width=0.95\textwidth]{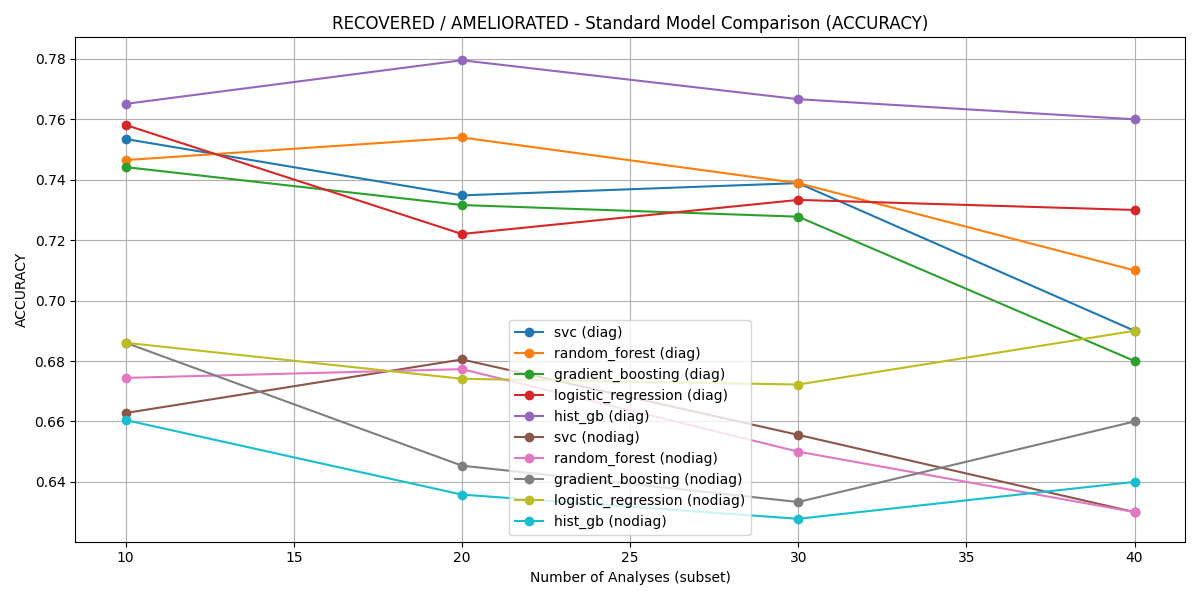}
\caption{Model accuracy comparison for Recovered vs.\ Ameliorated patients across diagnostic subsets.}
\label{fig:supp_task3_acc}
\end{figure}

\begin{figure}[ht]
\centering
\includegraphics[width=0.75\textwidth]{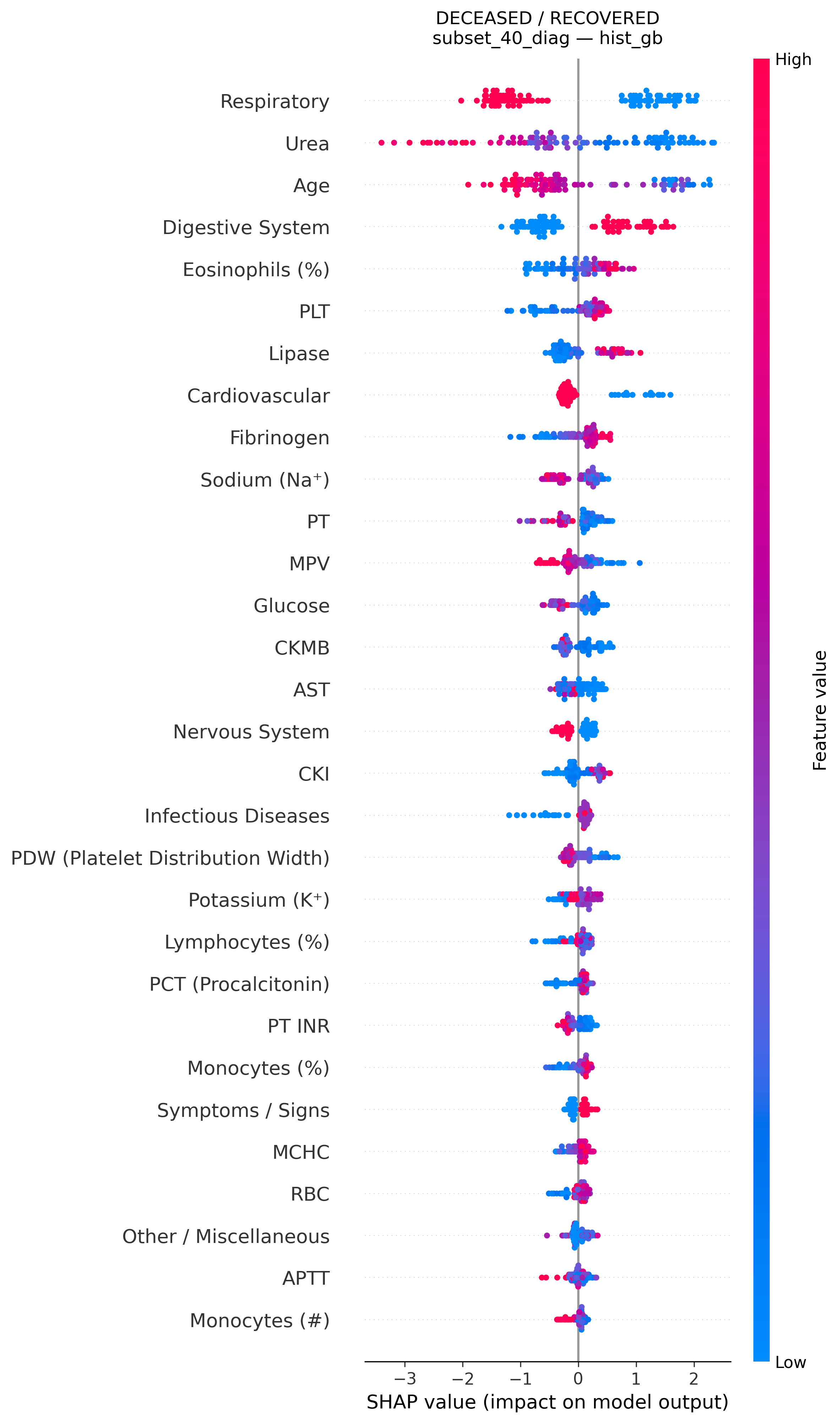}
\caption{SHAP summary plot for Task~2---Deceased vs.\ Recovered.}
\label{fig:supp_shap2}
\end{figure}

\begin{figure}[ht]
\centering
\includegraphics[width=0.75\textwidth]{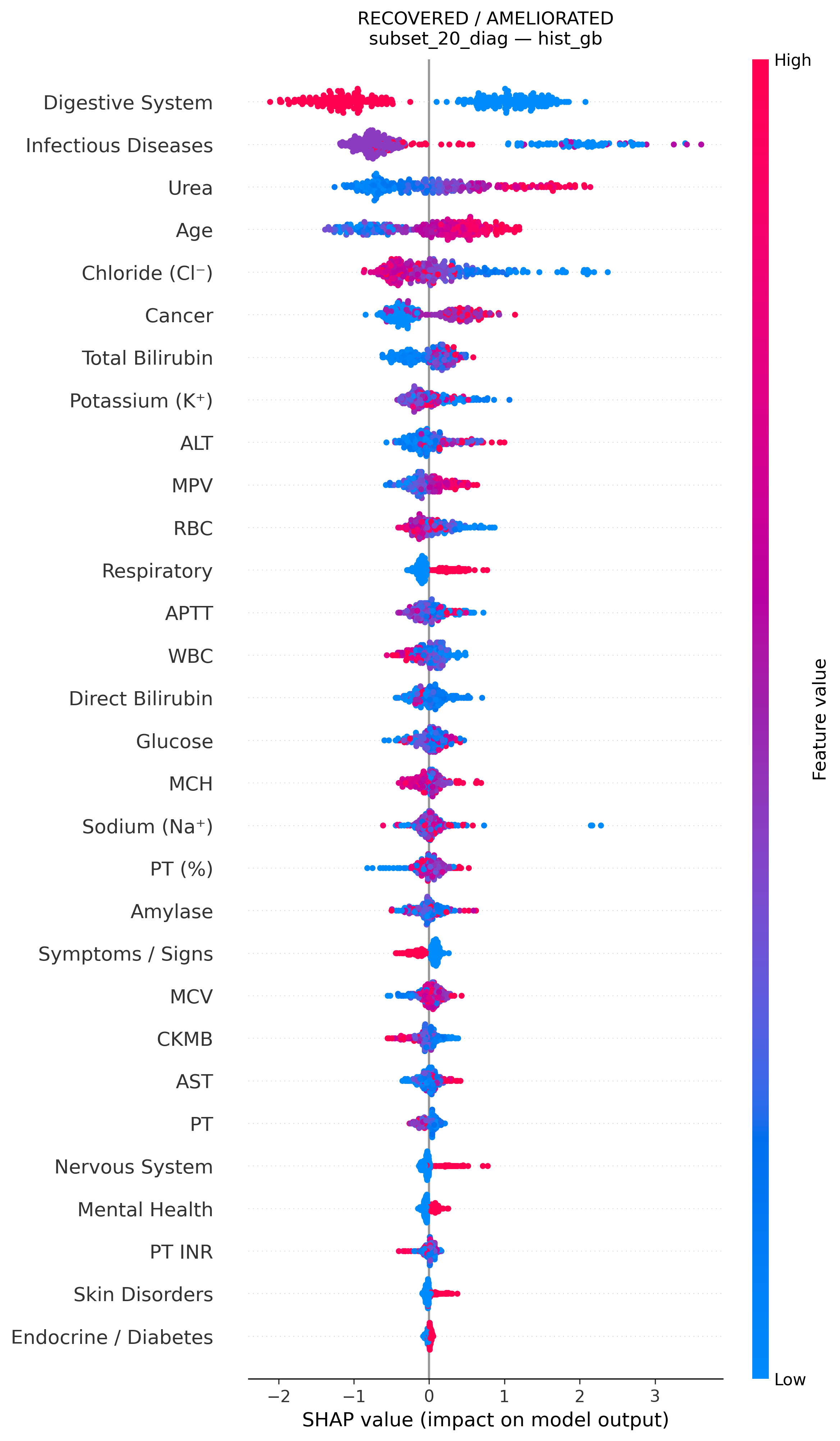}
\caption{SHAP summary plot for Task~3---Recovered vs.\ Ameliorated.}
\label{fig:supp_shap3}
\end{figure}


\clearpage
\begin{thebibliography}{32}

\bibitem{rudd2020}
Rudd KE, Johnson SC, Agesa KM, et al.
Global, regional, and national sepsis incidence and mortality, 1990--2017: analysis for the Global Burden of Disease Study.
\textit{The Lancet}. 2020;395(10219):200--211.
\href{https://doi.org/10.1016/S0140-6736(19)32989-7}{doi:10.1016/S0140-6736(19)32989-7}

\bibitem{acsqhc2021}
Australian Commission on Safety and Quality in Health Care.
\textit{A Review of the Impacts of Surviving Sepsis for Australian Patients}. 2021.

\bibitem{musat2025sepsis}
Mu\c{s}at F, P\u{a}duraru DN, Bolocan A, et al.
Sepsis Burden in a Major Romanian Emergency Center---An 18-Year Retrospective Analysis of Mortality and Risk Factors.
\textit{Medicina}. 2025;61(5):864.
\href{https://doi.org/10.3390/medicina61050864}{doi:10.3390/medicina61050864}

\bibitem{vincent1996}
Vincent JL, Moreno R, Takala J, et al.
The SOFA (Sepsis-related Organ Failure Assessment) score to describe organ dysfunction/failure.
\textit{Intensive Care Medicine}. 1996;22(7):707--710.
\href{https://doi.org/10.1007/s001340050156}{doi:10.1007/s001340050156}

\bibitem{rather2015}
Rather AR, Kasana B.
The Third International Consensus Definitions for Sepsis and Septic Shock (Sepsis-3).
\textit{JMS SKIMS}. 2015;18(2):162--164.
\href{https://doi.org/10.33883/jms.v18i2.269}{doi:10.33883/jms.v18i2.269}

\bibitem{donabedian2002}
Donabedian A.
The Apache II Severity of Disease Classification System.
\textit{An Introduction to Quality Assurance in Health Care}. 2002:159--162.
\href{https://doi.org/10.1093/oso/9780195158090.005.0005}{doi:10.1093/oso/9780195158090.005.0005}

\bibitem{legall1993}
Le Gall JR.
A New Simplified Acute Physiology Score (SAPS II) Based on a European/North American Multicenter Study.
\textit{JAMA}. 1993;270(24):2957.
\href{https://doi.org/10.1001/jama.1993.03510240069035}{doi:10.1001/jama.1993.03510240069035}

\bibitem{wongtangman2021}
Wongtangman K, Santer P, Wachtendorf LJ, et al.
Association of Sedation, Coma, and In-Hospital Mortality in Mechanically Ventilated Patients With Coronavirus Disease 2019--Related Acute Respiratory Distress Syndrome: A Retrospective Cohort Study.
\textit{Critical Care Medicine}. 2021;49(9):1524--1534.
\href{https://doi.org/10.1097/ccm.0000000000005053}{doi:10.1097/ccm.0000000000005053}

\bibitem{pollard2018}
Pollard TJ, Johnson AEW, Raffa JD, Celi LA, Mark RG, Badawi O.
The eICU Collaborative Research Database, a freely available multi-center database for critical care research.
\textit{Scientific Data}. 2018;5(1).
\href{https://doi.org/10.1038/sdata.2018.178}{doi:10.1038/sdata.2018.178}

\bibitem{chicco2020}
Chicco D, Jurman G.
Survival prediction of patients with sepsis from age, sex, and septic episode number alone.
\textit{Scientific Reports}. 2020;10(1).
\href{https://doi.org/10.1038/s41598-020-73558-3}{doi:10.1038/s41598-020-73558-3}

\bibitem{diwan2025}
Diwan S, Gandhi V, Baidya Kayal E, Khanna P, Mehndiratta A.
Explainable machine learning models for mortality prediction in patients with sepsis in tertiary care hospital ICU in low- to middle-income countries.
\textit{Intensive Care Medicine Experimental}. 2025;13(1).
\href{https://doi.org/10.1186/s40635-025-00765-5}{doi:10.1186/s40635-025-00765-5}

\bibitem{zhang2024}
Zhang G, Shao F, Yuan W, et al.
Predicting sepsis in-hospital mortality with machine learning: a multi-center study using clinical and inflammatory biomarkers.
\textit{European Journal of Medical Research}. 2024;29(1).
\href{https://doi.org/10.1186/s40001-024-01756-0}{doi:10.1186/s40001-024-01756-0}

\bibitem{bao2023}
Bao C, Deng F, Zhao S.
Machine-learning models for prediction of sepsis patients mortality.
\textit{Medicina Intensiva}. 2023;47(6):315--325.
\href{https://doi.org/10.1016/j.medin.2022.06.004}{doi:10.1016/j.medin.2022.06.004}

\bibitem{zeng2021}
Zeng Z, Yao S, Zheng J, Gong X.
Development and validation of a novel blending machine learning model for hospital mortality prediction in ICU patients with Sepsis.
\textit{BioData Mining}. 2021;14(1).
\href{https://doi.org/10.1186/s13040-021-00276-5}{doi:10.1186/s13040-021-00276-5}

\bibitem{hou2020}
Hou N, Li M, He L, et al.
Predicting 30-days mortality for MIMIC-III patients with sepsis-3: a machine learning approach using XGboost.
\textit{Journal of Translational Medicine}. 2020;18(1).
\href{https://doi.org/10.1186/s12967-020-02620-5}{doi:10.1186/s12967-020-02620-5}

\bibitem{brankovic2022}
Brankovic A, Hassanzadeh H, Good N, et al.
Explainable machine learning for real-time deterioration alert prediction to guide pre-emptive treatment.
\textit{Scientific Reports}. 2022;12(1).
\href{https://doi.org/10.1038/s41598-022-15877-1}{doi:10.1038/s41598-022-15877-1}

\bibitem{steitz2023}
Steitz BD, McCoy AB, Reese TJ, et al.
Development and Validation of a Machine Learning Algorithm Using Clinical Pages to Predict Imminent Clinical Deterioration.
\textit{Journal of General Internal Medicine}. 2023;39(1):27--35.
\href{https://doi.org/10.1007/s11606-023-08349-3}{doi:10.1007/s11606-023-08349-3}

\bibitem{yuan2025}
Yuan S, Yang Z, Li J, Wu C, Liu S.
AI-Powered early warning systems for clinical deterioration significantly improve patient outcomes: a meta-analysis.
\textit{BMC Medical Informatics and Decision Making}. 2025;25(1).
\href{https://doi.org/10.1186/s12911-025-03048-x}{doi:10.1186/s12911-025-03048-x}

\bibitem{akel2021}
Akel MA, Carey KA, Winslow CJ, Churpek MM, Edelson DP.
Less is more: Detecting clinical deterioration in the hospital with machine learning using only age, heart rate, and respiratory rate.
\textit{Resuscitation}. 2021;168:6--10.
\href{https://doi.org/10.1016/j.resuscitation.2021.08.024}{doi:10.1016/j.resuscitation.2021.08.024}

\bibitem{thiele2025}
Thiele D, Rodseth R, Friedland R, et al.
Machine Learning Models for the Early Real-Time Prediction of Deterioration in Intensive Care Units---A Novel Approach to the Early Identification of High-Risk Patients.
\textit{Journal of Clinical Medicine}. 2025;14(2):350.
\href{https://doi.org/10.3390/jcm14020350}{doi:10.3390/jcm14020350}

\bibitem{hu2022}
Hu C, Li L, Huang W, et al.
Interpretable Machine Learning for Early Prediction of Prognosis in Sepsis: A Discovery and Validation Study.
\textit{Infectious Diseases and Therapy}. 2022;11(3):1117--1132.
\href{https://doi.org/10.1007/s40121-022-00628-6}{doi:10.1007/s40121-022-00628-6}

\bibitem{he2024}
He B, Qiu Z.
Development and validation of an interpretable machine learning for mortality prediction in patients with sepsis.
\textit{Frontiers in Artificial Intelligence}. 2024;7.
\href{https://doi.org/10.3389/frai.2024.1348907}{doi:10.3389/frai.2024.1348907}

\bibitem{zhang2024lymph}
Zhang G, Wang T, An L, et al.
U-shaped correlation of lymphocyte count with all-cause hospital mortality in sepsis and septic shock patients: a MIMIC-IV and eICU-CRD database study.
\textit{International Journal of Emergency Medicine}. 2024;17(1).
\href{https://doi.org/10.1186/s12245-024-00682-6}{doi:10.1186/s12245-024-00682-6}

\bibitem{lin2024}
Lin TH, Chung HY, Jian MJ, et al.
AI-Driven Innovations for Early Sepsis Detection by Combining Predictive Accuracy With Blood Count Analysis in an Emergency Setting: Retrospective Study.
JMIR Publications Inc.; 2024.
\href{https://doi.org/10.2196/preprints.56155}{doi:10.2196/preprints.56155}

\bibitem{park2024}
Park SW, Yeo NY, Kang S, et al.
Early Prediction of Mortality for Septic Patients Visiting Emergency Room Based on Explainable Machine Learning: A Real-World Multicenter Study.
\textit{Journal of Korean Medical Science}. 2024;39(5).
\href{https://doi.org/10.3346/jkms.2024.39.e53}{doi:10.3346/jkms.2024.39.e53}

\bibitem{fan2024}
Fan SH, Pang MM, Si M, et al.
Quantitative changes in platelet count in response to different pathogens: an analysis of patients with sepsis in both retrospective and prospective cohorts.
\textit{Annals of Medicine}. 2024;56(1).
\href{https://doi.org/10.1080/07853890.2024.2405073}{doi:10.1080/07853890.2024.2405073}

\bibitem{li2025}
Li D, Hou J, Shi Z, et al.
Frailty Index-laboratory and lymphocyte subset patterns in predicting 28-day mortality among elderly sepsis patients: a multicenter observational cohort study.
\textit{Frontiers in Immunology}. 2025;16.
\href{https://doi.org/10.3389/fimmu.2025.1624655}{doi:10.3389/fimmu.2025.1624655}

\bibitem{choi2025}
Choi S, Nah S, Suh GJ, et al.
Prognostic Value of the AST/ALT Ratio in Patients with Septic Shock: A Prospective, Multicenter, Registry-Based Observational Study.
\textit{Diagnostics}. 2025;15(14):1773.
\href{https://doi.org/10.3390/diagnostics15141773}{doi:10.3390/diagnostics15141773}

\bibitem{pinte2025}
Pinte L, Dumitru AC, Usurelu AC, et al.
Low eosinophils and their dynamic as a predictor of death in patients with infections: a systematic review and meta-analysis of cohort studies.
\textit{Annals of Medicine}. 2025;57(1).
\href{https://doi.org/10.1080/07853890.2025.2541084}{doi:10.1080/07853890.2025.2541084}

\bibitem{abidi2008}
Abidi K, Khoudri I, Belayachi J, et al.
Eosinopenia is a reliable marker of sepsis on admission to medical intensive care units.
\textit{Critical Care}. 2008;12(2).
\href{https://doi.org/10.1186/cc6883}{doi:10.1186/cc6883}

\bibitem{shravani2025}
Shravani S, Kulkarni A, Aslam SM, Suhail KM, Shaji RM.
Absolute Eosinophil Counts as a Prognostic Marker in Patients with Sepsis.
\textit{Annals of African Medicine}. 2025;24(2):332--336.
\href{https://doi.org/10.4103/aam.aam_203_24}{doi:10.4103/aam.aam\_203\_24}

\bibitem{alduhailib2021}
Al Duhailib Z, Farooqi M, Piticaru J, Alhazzani W, Nair P.
The role of eosinophils in sepsis and acute respiratory distress syndrome: a scoping review.
\textit{Canadian Journal of Anesthesia}. 2021;68(5):715--726.
\href{https://doi.org/10.1007/s12630-021-01920-8}{doi:10.1007/s12630-021-01920-8}

\end{thebibliography}
\end{document}